# CNN Autoencoders for Hierarchical Feature Extraction and Fusion in Multi-sensor Human Activity Recognition

Saeed Arabzadeh[*], Farshad Almasganj[*], and Mohammad Mahdi Ahmadi[*]
*:Amirkabir University of Technology, Biomedical Engineering

*Abstract*— Deep learning methods have been widely used for Human Activity Recognition (HAR) using recorded signals from Inertial Measurement Units (IMUs) sensors that are installed on various parts of the human body. For this type of HAR, several challenges exist, the most significant of which is the analysis of multivarious IMU sensors data. Here, we introduce a Hierarchically Unsupervised Fusion (HUF) model designed to extract, and fuse features from IMU sensors data via a hybrid structure of Convolutional Neural Networks (CNN)s and Autoencoders (AE)s. First, we design a stack CNN-AE to embed short-time signals into sets of high dimensional features. Second, we develop another CNN-AE network to locally fuse the extracted features from each sensor unit. Finally, we unify all the sensor features through a third CNN-AE architecture as globally feature fusion to create a unique feature set. Additionally, we analyze the effects of varying the model hyperparameters. The best results are achieved with eight convolutional layers in each AE. Furthermore, it is determined that an overcomplete AE with 256 kernels in the code layer is suitable for feature extraction in the first block of the proposed HUF model; this number reduces to 64 in the last block of the model to customize the size of the applied features to the classifier. The tuned model is applied to the UCI-HAR, DaLiAc, and Parkinson's disease gait datasets, achieving the classification accuracies of 97%, 97%, and 88%, respectively, which are nearly 3% better compared to the state-of-the-art supervised methods.

## I. INTRODUCTION

Inertial Measurement Unit (IMU) sensors installed on different parts of the human body are widely used for Human Activity Recognition (HAR). This type of HAR, which for brevity we refer to it as HAR in this paper, stands out as a pivotal research area in healthcare and surveillance domains [1]. HAR can improve the diagnosis of a number of diseases or identify their severity, including Parkinson's and Alzheimer's disease [2]. The performance of HAR with IMU sensors is significantly enhanced when the sensors are placed on multiple body parts [3]. However, increasing the number of sensors complicates signal processing during feature extraction and feature fusion [4, 5]. To effectively enhance HAR performance, robust methods for analyzing and fusing data from multiple IMU sensors are essential [6].

To extract temporal and spatial features and improve the result of HAR, various Deep Neural Network (DNN) approaches, based on Convolutional Neural Networks (CNNs) and Recurrent Neural Networks (RNNs) have been proposed. Despite significant research efforts in DNNs for HAR, several fundamental challenges still persist. The major ones are: (1) the difficulty of data annotation, which hinders data collection, (2) the dependency of human actions on age, weight, and sex, and (3) the complexity of processing data obtained from multiple sensors.

To address the first challenge, unsupervised models such as Autoencoders (AE) and Boltzmann machines have been proposed [7, 8]. However, these approaches struggle with multi-sensor data and multi-class human activities.

Among the supervised methods, CNN-based architectures are a prominent approach for extracting human movement features within a framework of IMU sensors data [9]. These methods typically concatenate Multivarious IMU sensors Time Series (MITS) as input, and simultaneously perform feature extraction, feature fusion, and final classification [6, 10-14]. The performance of the CNN-based models often depends on the number of their layers and parameters. However, the limited amount of labeled data in HAR hinders the training of deep CNNs through supervised learning. Due to these limitations, such models are not always suitable for HAR purpose [8]. Consequently, alternative architectures, such as combining shallow CNNs with Long Short-Term Memory (LSTM) networks and attention mechanisms [8, 15, 16], have been proposed. While these models have improved HAR, their practical application remains limited, and typically relies on long input data durations [15, 17, 18].

Properly fusing MITS information could potentially address the aforementioned challenges in HAR and lead to improved activity recognition. Common fusion techniques include feature fusion, sensor fusion, and decision fusion, with feature fusion showing the most promising results [19]. In this work, we tackle HAR challenges by utilizing CNN and AE architectures to combine the feature extraction, feature fusion and sensor fusion tasks in a three-stage structure. We call this method Hierarchical Unsupervised Fusion (HUF) model. The output of the HUF is then applied to a final MLP classifier to accomplish the HAR task.

The HUF method is a hybrid model consisting of three cascaded blocks, resulting in the input signals being processed in three distinct steps. The first step involves applying independent AEs to each of the accelerometer and gyroscope signals, which produces an ensemble of individual features for every installed sensor unit.. In this step, each input signal is embedded to a high dimension feature set to capture its various informative aspects. This means the information from each signal is broken down into more detailed components to create a more meaningful representation of human movement for use in the subsequent steps. In the second step, we fuse the extracted features of the first step's network distinctively for each sensor unit. The aim is to create a salient pattern of activity from each sensor unit independently. This approach helps to prevent the fusion of features from incompatible MITS data combinations, particularly those features from non-activity periods [6, 8]. Finally, an additional AE network fuses the local features extracted from each sensor units. This step produces the final feature set, representing activity across various body movements.

The key contributions of this work are as follows.
- We introduce an end-to-end hybrid Neural Network (NN) architecture for efficient feature extraction and fusion from sensor axes to entire body sensors.
- We design a Stacked CNN-AE as a signal representative AE network to embed the short-time signal of each axis into high-dimensional features. These features efficiently represent and highlight various important aspects of the input signal, and decompose it into its distinct components.
- We propose a hierarchical feature fusion CNN-AE structure with two stages: the first stage fuses features within each sensor unit, while the second stage generates a unique feature vector from all sensors. This hierarchical approach enhances the model's ability to capture subtle patterns and variations, leading to improved recognition accuracy.
- Our approach is designed for recognizing activities using short time frames of input signals, enabling real-time HAR and making the system suitable for real-time applications.
- Our model utilizes a cascade of carefully designed autoencoder-based blocks, each of which is inherently robust against specific types of input noise or missing data.

The rest of this paper is organized as follows. Section II provides a brief overview of current studies in HAR. Section III describes the proposed NN-based method, detailing the functionalities of each component. Section IV covers the datasets used for evaluation, the training process for the model blocks, and the experimental results. Finally, Section V discusses the evaluation of the proposed method, and highlights the key findings and limitations of our work.

## II. RELATED WORKS

A typical DNN approach for HAR involves a CNN network followed by a Multi-layer Perceptron (MLP) classifier. In this structure, CNN primarily extracts local input patterns in its initial layers, and the global information fusion in its final layers. However, this network is sensitive to input noises and requires a large amount of labeled training data. Additionally, this model struggles with feature extraction and fusion on signals obtained from multiple sensors [6, 10-14, 20-22].

Several studies have investigated cascading CNNs with LSTM networks [8, 15, 16]. Although these models outperform simple CNNs, they require relatively long input signals. In other studies, attention mechanisms are integrated with CNN-LSTM structure to determine and focus on crucial features [15, 23]. In one work, a CNN model is empowered by three LSTMs and a spatial attention mechanism [23]. The reported results show that the model is sensitive to the input length variation and label annotation. Another approach utilizes dual-channel CNN-LSTM architectures to process data from different sides of the body [24, 25]. This approach aims to capture spatiotemporal features from each body sides and addresses the challenge of non-activity periods during analysis and fusion. However, its effectiveness relies on the availability of mirrored body sensor datasets.

To tackle challenges related to feature extraction and fusion, [26] proposes a CNN model that utilizes channel attention layers to enhance HAR performance. Unfortunately, the model's significant depth makes it practical only for extremely large datasets.

HAR methods using IMU sensors heavily rely on large datasets with meticulously labeled activity sequences. However, manual labeling process is resource-intensive, requiring significant human effort and computational power [7]. So, expanding the use of the unsupervised methods in this filed is highly appreciated [27, 28].

## III. METHODOLOGY

Our proposed HUF fusion model (Figure 1) processes multivariate time series data from multiple IMU sensors located on different parts of the body. Each sensor unit contains three accelerometers and three gyroscopes, in different x, y, and z axes, resulting in data structure of $s = [a_x(t), a_y(t), a_z(t), g_x(t), g_y(t), g_z(t)]$ for each unit. Therefore, for n sensor units, the input data is represented as:

$$X = [a_{x_1}(t), \quad a_{y_1}(t), \quad …, \quad g_{z_1}(t), \quad …, \quad g_{z_n}(t)]_{1 \times 6n} \quad (1)$$

where each element represents a time series signal. The proposed model involves four distinct processing steps, visualized in Figure 1. Each of these steps will be introduced individually in the subsequent parts of this section.

## A. Data Representation

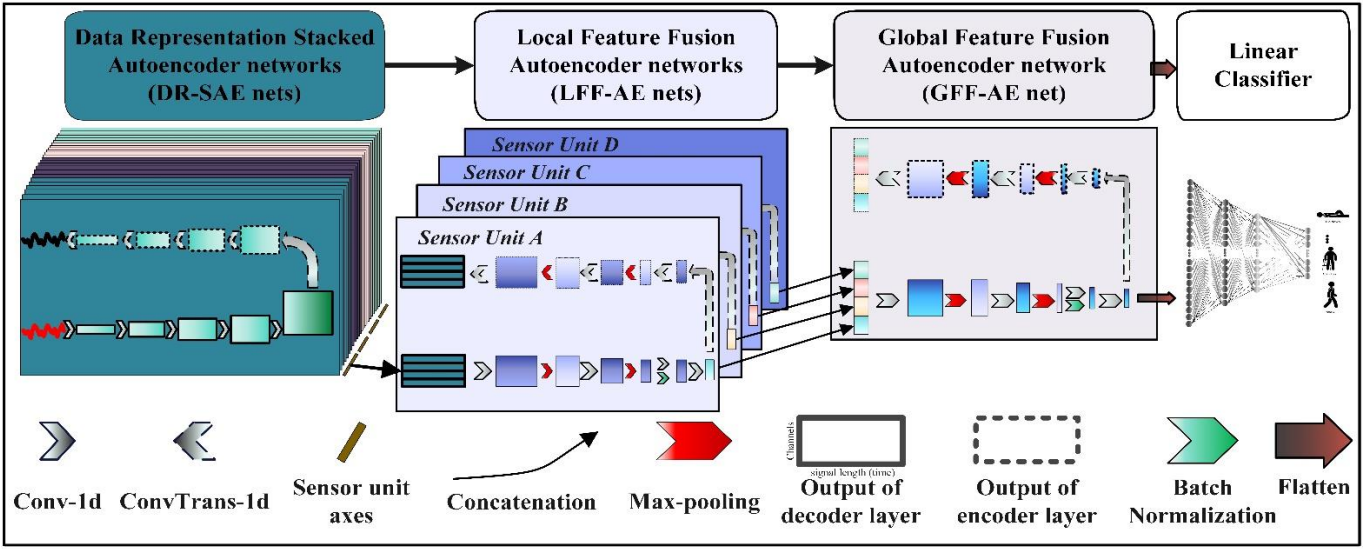

Figure 1: Architecture of the Proposed Hierarchically Unsupervised Fusion (HUF) Model with three steps of feature extraction and fusion with design CNN-AE.

To achieve a deep representation of the recorded signals of the accelerometers and gyroscopes, i.e., $x_i(t)$ $(i = 1, ..., 6n)$, we propose a stacked convolutional AE network, which is specifically designed to serve as an embedding step. The Data Representation Stacked (DR-SAE) architecture incorporates a number of convolutional layers within its encoder and decoder structures to hierarchically process the input time series segments (Figure 2). It generates a representation subspace with 256 views spanning different frequency bands of the input signal. The employed DR-SAE networks are trained using a stacked AE method, as illustrated in Figure 2. The training is done layer by layer, by freezing the previous layers and optimizer parameters. When all shallow architectures are cascaded, the resulted deep architecture captures the salience of signals in different scales; in this manner, it could decompose input signals with complex patterns, and extracts their hidden characteristics for further usage.

Obviously, an overcomplete AE architecture, like the designed DR-SAE, tends to learn the identity function in its hidden layers to copy the input to the code layer. This problem can be effectively mitigates by employing the stacked method, and freezing the trained layers by resetting optimizer parameters [29, 30]. This method forces the network to learn meaningful representations of the input signals, as illustrated in Figure 3. This figure depicts some samples of the features extracted in the DR-SAE code layer. It also displays a segment (the red box) of a raw input signal in Figure 3-d, alongside six samples of its extracted features, generated by some of the channels of the DR-SAE's code layer (Figure 3-a,b,c,e,f,g). These features (the pink lines) represent different aspects of the input signal that in many cases might be difficult to perceive directly from the raw signal. For example, Figure 3-a locates the input signal peaks and valleys.

The encoder section of the DR-SAE network can be treated as a nonlinear complex function, as given by: $F: x \to X, x \in \mathbb{R}^{H_0, W_0, C_0}, X \in \mathbb{R}^{H_5, W_5, C_5}$, where $H, W, C$ are the height, width and number of channels, respectively. It is obvious that $H_0 = C_0 = 1$, since the input signal is a time series. For simplicity, in layer $l$ of $F$, the $V_l = [v_{l1}, v_{l2}, ..., v_{lc}]$ is a set of filter kernels, where $v_{lc}$ refers to parameters that create the $c$-th channel output. The output of $F$ is given by

$$X = S\left(\sum_{k_5=1}^{C_5} S\left(\sum_{k_4=1}^{C_4} S\left(\sum_{k_3=1}^{C_3} S\left(\sum_{k_2=1}^{C_2} S\left(\sum_{k_1=1}^{C_1} x * v_{1k_1}\right) * v_{2k_2}\right) * v_{3k_3}\right) * v_{4k_4}\right) * v_{5k_5}\right) \quad (1)$$

where * denotes the convolution operator, $S$ is the SELU activation function, and each parenthesis indicates the output of the layer that is equipped with $c_i$ filters. The SELU function is a proper activation function for HAR tasks because the decoder

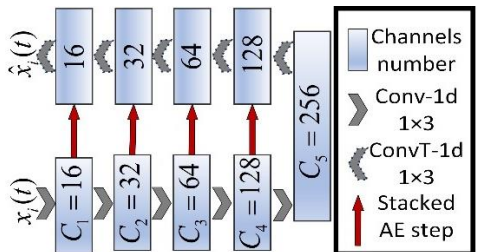

Figure 2: Architecture of the Data Representation Stacked Autoencoder (DR-SAE) – Red arrows indicate the training steps for each stacked AE layer within the DR-SAE.

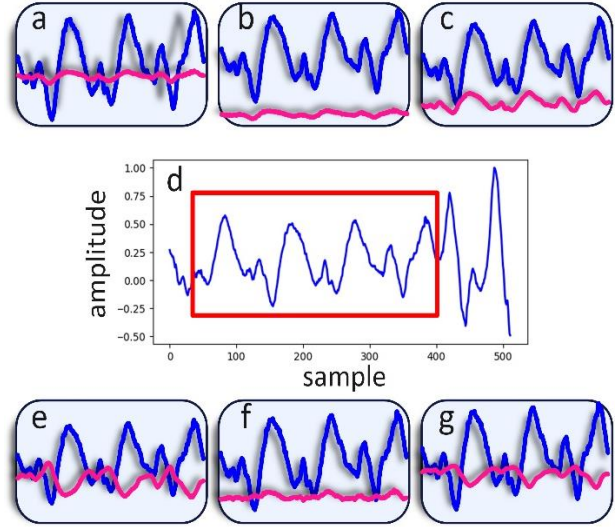

Figure 3: The d is an input window of DR-SAE net that creates the pink line representing in a, b, c, e, f and g in its code layer. To compare the created signal in the code layer, a cropped portion of raw signal (blue line) is plotted with the created signals

aims to reconstruct the input signal in its output. Since some parts of the input signal are negative, it is essential to these negative parts. This function is approximately linear for input values within a specific range $(-1, 1)$. By accepting this assumption that the activation function acts approximately linear in most of the activities of the employed convolutional layers, the output of each channel of a convolutional layer can be formulated by a summation through all previous channels that implicitly entangles with the spatial correlation [31]. For example, if we consider the output of the first channel of the second layer, it can be formulated by:

$$X_{21} = (x * v_{11}) * v_{21_1} + (x * v_{12}) * v_{21_2} + (x * v_{13}) * v_{21_3} + \cdots + (x * v_{1c_1}) * v_{21_{c_1}} \quad (2)$$

where $v_{21_i}$ shows the $i$-th filter in the first channel of the second layer, convolved with the $i$-th output from the first layer. This can be further expressed as:

$$X_{21} = x * (v_{11} * v_{21_1} + v_{12} * v_{21_2} + v_{13} * v_{21_3} + \cdots + v_{1c_1} * v_{21_{c_1}}) \quad (3)$$

This formulation explains that each output of the second layer is made by the summation of a bundle of FIR filters, such as $v_{11} * v_{21_1}, \ldots, v_{1c_1} * v_{21_{c_1}}$. If we extend the formulation up to the code layer, the summations will involve higher-degree FIR filters, for instance $v_{11} * v_{21_1} * v_{31_1} * v_{41_1} * v_{51_1}$. Figure 4 shows the frequency response of a few of these filters by only changing the involved code layer kernels after the training phase of the network. It is evident that the frequency responses of the resulting filters are noticeably different which in turn, discriminate and represent various aspects of the input signal in the code layer outputs.

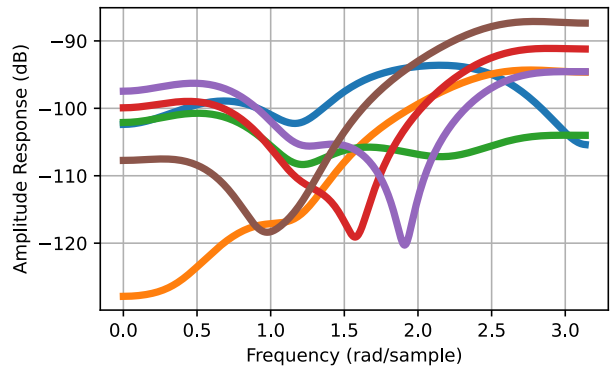

Figure 4: Frequency response of some randomly selected filters that form the output of the code layer in DR-SAE.

## B. Feature fusion

### 1) Local Feature Fusion

As shown in Figure 1, the extracted features by the DR-SAEs are firstly stacked in a tensor for each sensor unit. Next, a Local Feature Fusion Autoencoder network (LFF-AE) is proposed to implement the fusion process. As shown in Figure 5, six sets of features extracted from the accelerometers and gyroscopes signals of a sensor unit are concatenated and feed the input of an LFF-AE. By passing through four convolutional layers of the encoder of LFF-AE, the fused version of the input features is available as $C_4$ output channels. In this manner, multiple salient patterns of different sensor units, which are mounted in different parts of the body, are now available and ready to be fused in an additional network in the third step of the proposed model.

To further reduce the size of the extracted fused features and enhance their robustness to input variations, two max pooling layers are put after the first two convolutional layers, respectively. As reported in [32], this architecture results in lower sensitivity of the features to slight distortions and variations within the input data. Additionally, the batch normalization layer (BN) that adjust the amplitude of the code layer output is embedded just after the third convolutional layer, which experimentally is verified that leads to better final classification results. The BN layer works over the output of the previous layer channels, by calculating and using the mean and variance over the input mini-batches, during the network training phase [33]. As depicted in Figure 5, the decoder and encoder together form a completely symmetric structure, with the exception that the encoder contains an extra BN layer. All convolutional and trans-convolutional layers utilize a kernel size of $1 \times 3$. The max-pooling filters are correspondingly $1 \times 4$ and $1 \times 3$, with stride two. Moreover, different values are explored for the channels number ($C_1$ to $C_4$) to identify the optimal settings that lead to the best classification performance. The process of tuning the hyperparameters is detailed in Section IV.C.

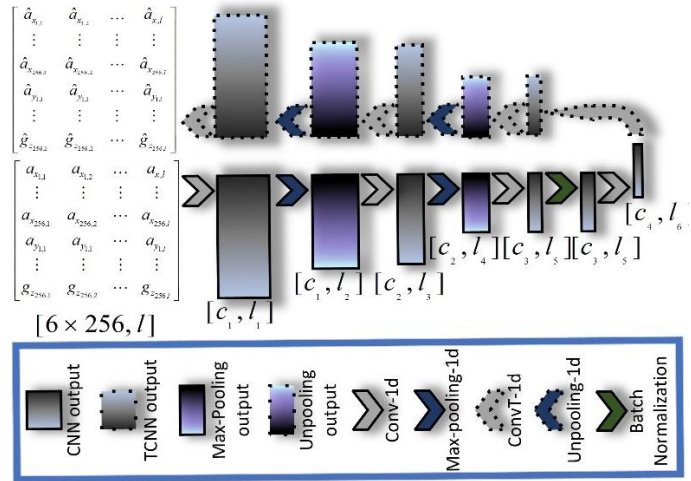

Figure 5: Locally Feature Fusion-AE architecture – It takes all features outputs of DR-SAE nets with same sensor unit with length l to combine them into $c_4$ channels with length $l_6$.

### 2) Global Feature Fusion

The third step of the proposed approach involves fusing all the feature sets extracted up to this step to obtain a unique set of richly combined features from different movements of the body parts. This step is performed by designing and employing the Global Feature Fusion AutoEncoder (GFF-AE). The finally resulted features extensively represent the outputs of all the *n* sensor units mounted on different parts of the body. By using the proposed hierarchical approach, we prevent the premature fusion of different input features during movement recognition. For example, in hand-related activities, the contribution of foot-mounted sensors in recognizing the movement is negligible; therefore, combining its outputs with other sensors outputs, in one step, may mislead the final classifier network; the proposed multi-step fusion structure successfully overcomes this weakness.

The input of GFF-AE is a tensor that contains all the outputs from the code layers of the former LFF-AEs. This combined representation reflects the effects of the current activities of all sensors installed on different positions of the body. The architecture of GFF-AE net is similar to LFF-AE (Figure 5) except in the details of reducing the dimensions of $C_1$ to $C_4$ layers, which is discussed in Section IV.C.

## C. Classification

We employ a fully connected network with four subsequent layers as a classifier to evaluate the proposed hierarchical feature extraction and feature fusion method. Obviously, the numbers of neurons in its different layers are varied according to the number of the output features extracted via the proposed HUF model, and the number of activities that must be investigated and recognized.

## IV. EXPERIMENTS

### A. Datasets

We employ three datasets to evaluate our proposed HUF model, and address various challenges in HAR: (1) the **Daily Life Activities** (DaLiAc) dataset [17] with 13 HA tasks, recorded by using a complex setup with many sensors mounted on different parts of the body; (2) the **UCI HAR** dataset [34], recorded by employing a smartphone placed on the waist, representing a dataset with a single sensor; and (3) the **Multimodal Dataset of Freezing of Gait in Parkinson's Disease** (MDFG-PD) dataset, focusing on Parkinson's disease (PD) gait with various age and PD ranges. To record MDFG-PD dataset, different devices were employed which we only focus on IMU sensors to detect the walking with or without Freezing of Gait (FOG). MDFG-PD poses additional challenges due to sensor disconnections during walking [35], which requires a robust method to fuse the features of the involved sensors. Table 1 and Figure 7 provide additional insights into the introduced datasets and the distribution of each of the tasks.

### B. Segmentation

Sliding window segmentation offers advantages for time series analysis, particularly in identifying localized features like short-term trends, and also in improving computational efficiency for large datasets [36]. However, it presents a challenge for HAR researches that focus on short-time activity recognition. For example, the small segments might not perfectly capture the start and end points of activities, and this imposes a difficult task to the network to learn many considerable variations occurred in the segments boundaries. In order to deal with this problem by increasing the training data, and enhancing the robustness of the trained networks, we employ a 50% overlap between adjacent sliding windows. Moreover, we split data into training and test sets, using a subject-based hold-out strategy (70% for training and 30% for test); the data is shuffled inside the training partition. The purpose of applying these techniques is to increase the network's ability to detect activities in short time frames to make it ready for handling real-world scenarios.

*Table 1: Details of the employed datasets. ACC: Accelerometer, Gyro: Gyroscope*

| Dataset | Participants | sensor model | Num of sensor | Placed sensors | Activities | Sample frequency | Window size | Window overlap | Age (mean±std) |
|---|---|---|---|---|---|---|---|---|---|
| DaLiAc | 19 | ACC Gyro | 4 | Right hip Chest Right wrist Left ankle | Sitting, Lying, Standing, washing dishes, vacuuming, sweeping, walking, ascending stairs, descending stairs, bicycling on ergometer (50 and 100 W), Rope jumping | 200 Hz | 512 | 50% | $26 \pm 8$ |
| UCI-HAR | 30 | ACC Gyro | 1 (Cell phone) | Waist | standing, sitting, laying down, walking, walking downstairs, walking upstairs | 50 Hz (Up sampling to 100 Hz) | 256 | 50% | 19-48 |
| MDFG-PD | 12 | ACC Gyro | 4 | Arm, Waist, Right shank, Left Shank | Walking forward, walking with obstacle, walking with turning 90° and 360°, walking at high speed. (The aim is finding FOG during different kinds of walking) | 100Hz (Up sampling to 500 Hz) | 1500 | 50% | 69.1 $\pm$ 7.9 |

### C. Experimental Settings

#### 1) Model structure

The structure of the proposed model applied to each of the mentioned databases should be adapted to the specifications of that database. For the DaLiAc dataset, the structure is exactly as shown in Figure 1. For the UCI HAR dataset, since it is recorded by employing a single sensor unit in a mobile phone, we can omit the GFF-AE part whose task is the fusion of the output data of multiple sensors; therefore, it utilizes only one LFF-AE net for feature fusion. For the case of the MDFG-PD dataset, the architecture remains the same as shown in Figure 1. In this dataset, for some of the subjects, some sensor units are inactive. When this problem occurs, all blocks of the proposed model, except for the GFF-AE block, continue to function normally because the DR-SAE and LFF-AE blocks rely only on the active sensors. However, in the GFF-AE block, the inputs from the missing sensors are set to zero during both the training and test phases. In this manner, the deactivation of a sensor only affects a small part of the GFF-AE block's input, which is only one part of the entire HUF model.

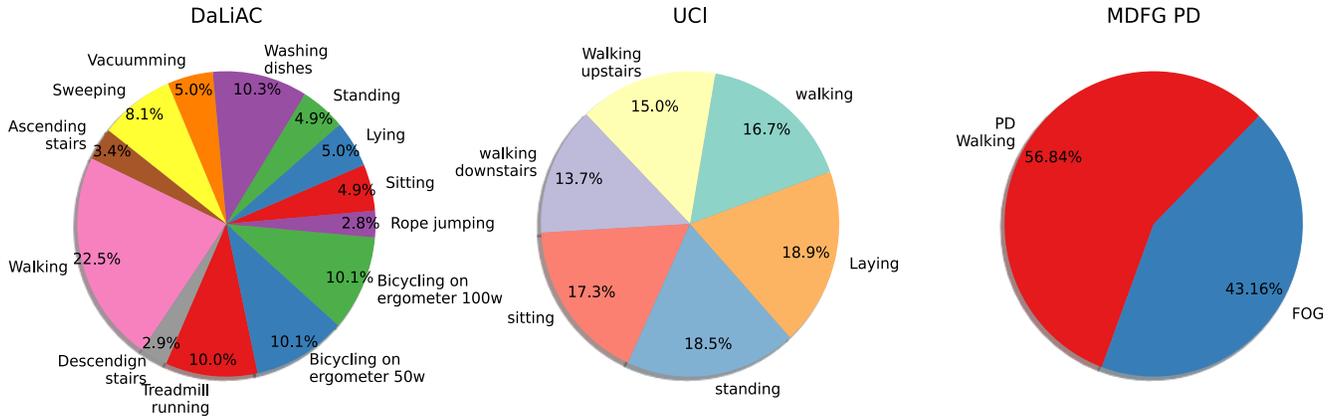

Figure 7: Class population in three different datasets.

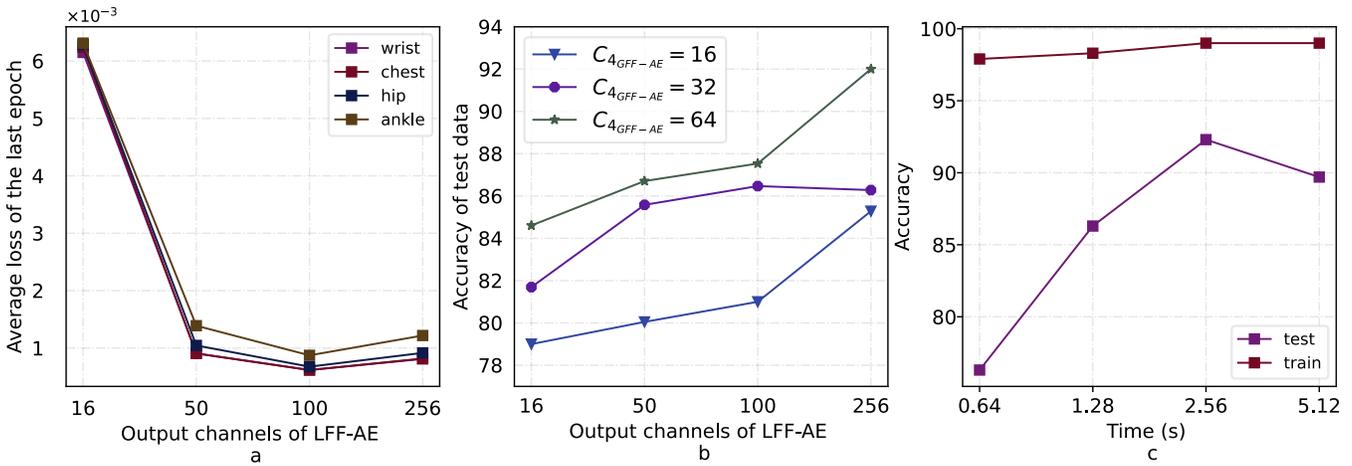

Figure 7: a) Average loss of LFF-AE versus the number of the code layer kernels (in the last training epoch). B) Accuracy of the final classifier due to the code layer sizes of the LFF-AEs and GFF-AE. C)Accuracy of the final classifier versus the input time length for the best hyperparameters of LFF-AEs and GFF-AE.

In the following sections, we will analyze the test results to evaluate the robustness of the proposed model against the data loss of certain sensors.

2) *Hyperparameters Configuration*

   *a) DR-SAE Network*

The hyperparameters of this network are empirically chosen as depicted in Figure 2, and are the same for all the employed datasets. This network is an overcomplete AE, without incorporating max-pooling layers. Furthermore, the activation function of CNN layers is the SELU function that has self-normalizing properties, providing a high-level robustness during the learning process. Moreover it allows better training of networks with many layers [37]. In the following, we examine different kernel sizes such as $(1 \times 3), (1 \times 5)$ and $(1 \times 7)$. We found that kernel size of $(1 \times 3)$ leads to better results in the evaluation phases. The training phase of a network continues until the loss function becomes less than 0.005, and at least five epochs pass.

   *b) LFF-AE and GFF-AE networks:*

To evaluate and verify the effectiveness of various feature reduction steps taking place inside the proposed HUF model, we conducted some designed experiments on the DaLiAc dataset. This dataset includes data from four IMU sensors mounted on four different parts of the body. For each sensor unit, an assigned LFF-AE recieves six distict feature sets, extracted from three accelerometers and three gyroscops outputs, as detailed in Figure 5. Every feature set consists of 256 feature vectors with the length of $l$, which concatenate to form a matrix with size $(6 \times 256, l)$ as a LFF-AE network block input. This input is transformed into a matrix with the size of $(C_{4_{LFF-AE}}, l_6)$ in the code layer where $C_{4_{LFF-AE}}$ represents the number of code layer kernels, and $l_6$ is the channel length of the generated code. Next, the resulting codes from four exploited LFF-AE blocks are concatenated to form the input of the subsequent GFF-AE network as depicted in Figure 1.

To reduce the number of extracted features during the fusion process, we can reduce the number of kernels in the code layers of LFF-AE and GFF-AE blocks. To assess the influence of the number of kernels in LFF-AE code layer ($C_{4_{LFF-AE}}$) in the training phase, and its impact on the subsequent steps, we searched for the best value of this hyperparameter among the four different numbers of 16, 50, 100, and 256. As we shown in Figure 7-a, the LFF-AE network achieved good training performance when $C_{4_{LFF-AE}}$ was chosen 50 or larger. Additionally, for each of these four code sizes, we examined three different fusion settings in the GFF-AE, using kernel sizes ($C_{4_{GFF-AE}}$) of 16, 32, and 64. This allowed us to investigate the effect of kernel size during the final feature fusion stage and its influence on the final classifier performance. As shown in Figure 7-b, a smaller number of features extracted in the LFF-AE ($C_{4_{LFF-AE}}$) negatively impacts the training of the subsequent GFF-AE and consequently, degrades the classifier's results. The best performance was obtained by setting $C_{4_{LFF-AE}}$ to 256, and $C_{4_{GFF-AE}}$ to 64. These values were selected and fixed for the HUF model, and were used for all the three datasets.

3) *input size*

The size of the input data is another important parameter that greatly impacts the HAR performance. While, the HAR models with short length inputs are more practical, this hyperparameter has often been ignored in previous studies. Figure 7-C illustrates the effects of the input length on the obtained HAR accuracies, while applying HUF model to the DaLiAc dataset. A proper input pattern length is investigated in the range of 0.64 to 5.12 seconds, and the obtained accuracy results, using the proposed HUF model, are shown in Figure 7-c. The figure shows that increasing the input length typically leads to higher test accuracies. However, a slight decrease in the recognition accuracy occurs on the test data for the input duration of 5.12 seconds. This decrease happens due to a trade-off between input pattern length and the available training input data segments. As the input segment length increases, the number of the available training segments significantly decreases. This potentially leads to the model overfitting, and a wider gap between the accuracy results obtained over the training and test data.

## D. Results

This section explores the benefits of our proposed feature extraction and fusion method against some of the previous state-of-the-art methods that focused on supervised learning methods. A key advantage of our proposed model is its inherent robustness during feature extraction and fusion, which is especially valuable when dealing with conditions that the data of a sensor is lost during the recording phase. The inherent adaptability of the proposed unsupervised method allows it to handle variations and challenges in a non-annotated data, and makes it well-suited for real-world applications.

Table 2 presents the experimental results obtained over DaLiAc and UCI-HAR datasets by the feature extraction, feature fusion and classification via the proposed method, alongside some other state-of-the-art methods. It is noteworthy that our results are obtained from relatively short-time inputs, and are superior to most of the recent studies. Of course, there are two exceptions in Table 2: Debache et. al. [38] method in which some handcrafted features are utilized, or Huynh et. al. [39] method in which the developed model needs to a long-time input data extracted from image-transformed data.

*Table 2: Results of the proposed Unsupervised method against some state-of-the-art supervised methods - IT: Input Time-length (s) – Acc: Accuracy*

| | **Employed Model** | **IT(s)** | **Acc(%)** |
|---|---|---|---|
| **DaLiAc** | **LSTM** [40] | 64 ∗ 0.6 | 82.5 |
| | **CNN** and handcraft features [38] | 5 | 95.4 |
| | **AcImgEncoding + CNNs** [39, 41] | 5 | 95.7 |
| | **SVM, AdaBoost, KNN** [17] | 10 | 89 |
| | **Proposed Method** | 2.56 | 92.9 |
| **UCI- HAR** | **LSTM** [18] | 10 | 89 |
| | **MarNASNet** – Mobile aware CNN Sensor-based [42] | 2.56 | 92. |
| | **CNN** [43] | 2.8 | 90 |
| | **CNN-LSTM** (Series) [43] | 2.8 | 90 |
| | **CNN-LSTM** (Paraller) [43] | 2.8 | 90 |
| | **Deep CNN-LSTM with attention** [15] | 40.9 | 94 |
| | **Proposed Method** | 2.56 | 96.8 |

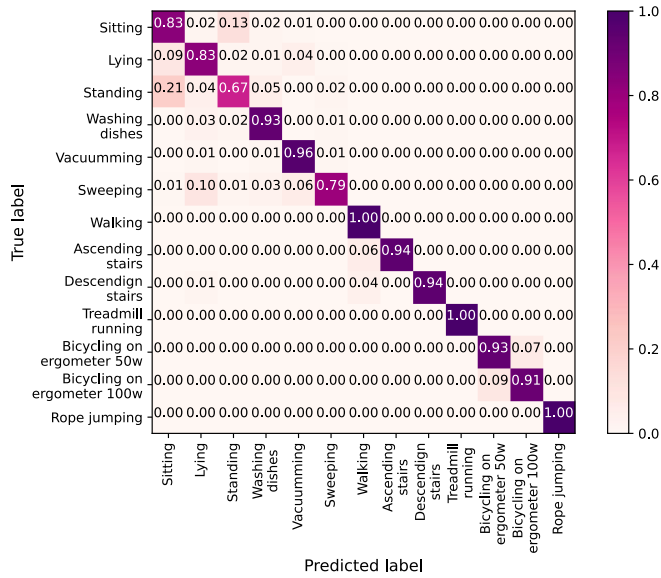

*Figure 8: Confusion matrix of the proposed model when applied to the DaLiAc dataset*

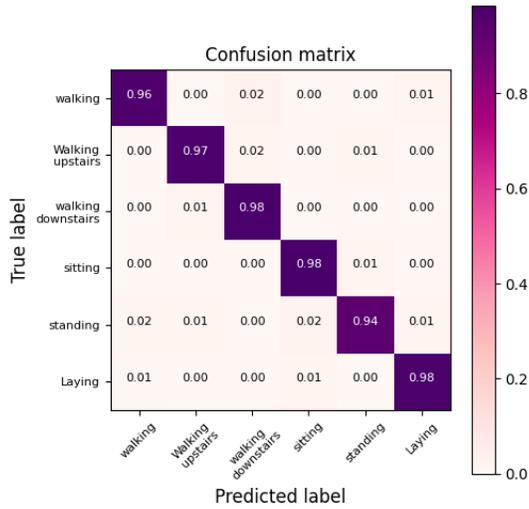

*Figure 9: Confusion matrix of the proposed model obtained for the UCI HAR dataset.*

For a more in-depth analysis, Figure 8 shows the confusion matrix of the proposed method applied to DaLiAc dataset. It can be seen that activities such as sitting and standing are prone to misclassification, especially due to their very similar static postures.

The second part of Table 2 presents the classification result of our proposed method over the UCI-HAR dataset against some other state-of-the-arts methods. As can be seen, the proposed HUF model performs well on this smartphone dataset, even though it operates in an unsupervised manner, and receives short-time input signals. The confusion matrix depicted in Figure 9 reveals that the highest misclassification rate occurs for sitting and standing activities, which can be attributed to their similar static nature as detected from the outputs of only one waist-mounted sensor.

Finally, the proposed model is evaluated in recognizing the FOG (freezing of gait) in the PD patients, using the MDFG PD dataset. Detecting this abnormality during continuous walking is typically a difficult task. The proposed feature confusion model aims to identify the FOG by recognizing gait steps that deviate from normal walking. Table 3 shows the FOG classification results, on the MDFG PD database, obtained by the proposed method, a traditional SVM-based method [44], and a CNN-based state-of-the-art method [45]. It is important to note that [45] reports the accuracy while the data from subjects with faulty missed sensors were excluded from the dataset. The superiority of our method over the other two was achieved while the subjects with missing sensors were not excluded from the training and test datasets.

*Table 3: FOG recognition results obtained by the proposed method and two other state-of-the-art methods, using the MDFG PD database.*

| Method | SVM (Handcraft features) [44] | IMUFoGNET (five CNN layers + three dense layers) [45] | HUF method features with MLP classifier. |
| --- | --- | --- | --- |
| Accuracy | 85.5 | 80.4 | 87.6 |

## V. CONCLUSIONS

In this article, we presented a professionally designed deep model that greatly benefits from the hybrid architecture of CNN-based autoencoders for application to the HAR task. The proposed method specifically tackles the challenges associated with feature fusion from various sensors in scenarios involving weakly labeled datasets and multiple IMU sensors. The first key aspect of our approach lies in embedding signals from individual axes of each sensor, extracting extended beneficial features from them. This allows us to effectively fuse the individual extracted feature sets in the subsequent two steps, ultimately enabling the HUF model to represent activities across the entire body. This approach offers a valuable solution for utilizing unsupervised trainable models in HAR tasks, particularly when dealing with datasets that lack extensive labeled data. Moreover, due to its hierarchical architecture, the model demonstrates noticeable robustness against noisy input data and missing sensor conditions.

The proposed model demonstrates some limitations in differentiating highly similar static activities such as sitting, lying, and standing, as depicted in Figure 8, for the DaLiAc dataset. This can be attributed to the inherent closeness in movement patterns between the short time activities. However, when evaluated on the UCI-HAR dataset, the model achieves superior accuracy compared to the previous models. This is particularly evident in its ability to recognize walking activities and static postures with greater precision. Furthermore, the model exhibits robust performance when applied to the MDFG-PD dataset. This dataset presents an additional challenge due to the presence of missing sensor data in some subjects. The effectiveness of the model in handling such data underscores its ability to perform well in real-world scenarios where data imperfections are often inevitable.

Training the proposed HUF hybrid network is computationally expensive, especially for some employed stacked AEs, but this computational cost is not a major issue as it can be greatly minimized during the deployment of the fully trained model.